\documentclass[letterpaper, 10 pt, conference]{ieeeconf}

\usepackage{graphicx}
\usepackage{adjustbox}
\usepackage{wrapfig}
\usepackage[labelfont=bf,figurename=Fig.,font=small]{caption}
\usepackage{subcaption}
\usepackage{lipsum}
\usepackage{amssymb}
\usepackage{amsmath}
\usepackage{siunitx}
\usepackage{mathtools}
\usepackage{xcolor}
\usepackage{hyperref}
\usepackage[ruled,vlined,linesnumbered]{algorithm2e}
\usepackage[noend]{algpseudocode}
\usepackage{ifthen}
\usepackage{dsfont}
\usepackage[backend=biber,style=numeric-comp,sorting=none,maxbibnames=99]{biblatex}
\usepackage{balance}
\usepackage{dirtytalk}
\usepackage{afterpage}

\usepackage{times}

\allowdisplaybreaks

\newcommand{\ra}{\rightarrow}
\newcommand{\R}{\mathbb{R}}

\newtheorem{proposition}{Proposition}
\newtheorem{definition}{Definition}
\newtheorem{lemma}{Lemma}

\newtheorem{corollary}{Corollary}

\newcommand{\interplayermeas}{\overline{z}}
\newcommand{\landmarkmeas}{z}

\bibliography{references}

\IEEEoverridecommandlockouts
\title{\LARGE \bf GTP-SLAM: Game-Theoretic Priors for\\ Simultaneous Localization and Mapping in Multi-Agent Scenarios
}

\author{Chih-Yuan Chiu$^{1}$ and David Fridovich-Keil$^{2}$
\thanks{$^{1}$Department of Electrical Engineering and Computer Sciences, University of California, Berkeley, CA 94720.
\href{mailto:chihyuan_chiu@berkeley.edu}{\tt\small chihyuan\_chiu at berkeley dot edu}.}%
\thanks{$^{2}$Department of Aerospace Engineering, The University of Texas at Austin, Austin, TX, 78712, USA
\href{mailto:dfk@utexas.edu}{\tt\small dfk at utexas dot edu}.}%
}

\begin{document}

\maketitle

\thispagestyle{empty}
\pagestyle{empty}





\begin{abstract}

Robots operating in multi-player settings must simultaneously model the environment and the behavior of human or robotic agents who share that environment. This modeling is often approached using Simultaneous Localization and Mapping (SLAM); however, SLAM algorithms usually neglect multi-player interactions. In contrast, the motion planning literature often uses dynamic game theory to explicitly model noncooperative interactions of multiple agents in a known   environment with perfect localization. Here, we present GTP-SLAM, a novel, iterative best response-based SLAM algorithm that accurately performs state localization and map reconstruction, while using \emph{game theoretic priors} to capture the inherent non-cooperative interactions among multiple agents in an uncharted scene. By formulating the underlying SLAM problem as a potential game, we inherit a strong convergence guarantee. Empirical results indicate that, when deployed in a realistic traffic simulation, our approach performs localization and mapping more accurately than a standard bundle adjustment algorithm across a wide range of noise levels.

\end{abstract}

\section{INTRODUCTION}
\label{sec: Introduction}

To navigate safely and efficiently in real-world scenarios, autonomous vehicles must accurately represent dynamic, uncharted environments, and execute robust motion plans that account for multi-player interactions in the scene. This requires a careful fusion of state estimation, prediction, and path planning modules in vehicle autonomy stacks. However, current, state-of-the-art autonomous navigation pipelines typically treat these problems separately and do not incorporate direct feedback between them.

Regarding estimation, algorithms tackling the Simultaneous Localization and Mapping (SLAM) problem aim to accurately reconstruct an uncharted environment while also localizing the ``ego'' player within it \cite{LeonardDurrantWhyte1991SimultaneousMapBuilding, Cadena2016PastPresentandFuture, davison2018futuremapping1, davison2019futuremapping2}. In recent years, popular inference algorithms for SLAM, such as factor graph-based methods, have also been used to solve challenging motion planning and optimal control problems jointly with SLAM problems in unknown environments \cite{Dellaert2017FactorGraphsforRobotPerception}.
However, these approaches often do not account for purposeful interactions among dynamic players in the environment. 
It is thus unclear whether these methods can truly model safe, efficient, and robust motion plans generated by an ego player operating in the vicinity of other players.

Interactions between independently-minded, mobile players are naturally modeled as dynamic games between rational actors with differing objectives \cite{isaacs1954differential, basar1998DynamicNoncooperativeGameTheory, starr1969nonzero, starr1969further}.
Recent advances in game-theoretic motion planning exploit this structure to predict the responses of other players to one's own decisions, and identify a desirable equilibrium strategy \cite{fisac2015reach, fridovich2019efficient}. To the best of the authors’ knowledge, however, game-theoretic formulations of noncooperative multi-player interactions have not yet been considered in SLAM tasks.


In this work, we formulate the SLAM task from the perspective of an ego player, who is interacting with multiple other players while simultaneously estimating all players' positions and all landmark locations. Inspired by the dynamic game theory literature, we first establish mild assumptions under which this problem can be formulated as a potential game. 
We then present a factor graph-based algorithm to solve this game and prove that it is guaranteed to converge to a local equilibrium. 
Unlike existing SLAM methods, this approach tightly integrates estimation, prediction, and decision-making \emph{for multiple players, simultaneously}. 
Empirical results illustrate that, compared to standard bundle adjustment, incorporating game-theoretic interaction priors leads to higher localization and map reconstruction accuracy in a realistic traffic scenario.




\section{RELATED WORK}
\label{sec: Related Work}

\subsection{Simultaneous Localization and Mapping}

Simultaneous Localization and Mapping is a fundamental state estimation task with a well-developed literature in the robotics community \cite{LeonardDurrantWhyte1991SimultaneousMapBuilding, Cadena2016PastPresentandFuture}, the unsolved aspects of which continue to attract great interest \cite{davison2018futuremapping1, davison2019futuremapping2}. A standard method for solving SLAM problems is to reformulate the underlying maximum a posteriori (MAP) estimation problem into a nonlinear least squares problem, which can then be solved via factor graph optimization \cite{DellaertKaess2006SquareRootSAM,  KaessDellaert2012iSAM2}.

In recent years, factor graphs have been used to formulate a wide range of robotics problems beyond the SLAM task in static environments, including model predictive control and trajectory tracking \cite{TaDellaert2014AFactorGraphApproachtoEstimationAndMPC, Dellaert2017FactorGraphsforRobotPerception}. Factor graph-based methods have also been used to solve the dynamic SLAM problem, which involves the reconstruction of uncharted environments with dynamic players \cite{ zhang2020vdoslam} who share measurement information, and perform estimation, and prediction in a cooperative game framework \cite{ZhangDellaert2021MRiSAM2, Hua2013NewAlgorithmMergingGameIntoEKF}. These methods typically infer time-dependent variables pertaining to multiple players without accounting for players' interactive, and likely noncooperative, behavior \cite{Ke2021EKFBasedGameforCooperativeWirelessNetworkNavigation, Tang2020GameTheoryCooperativeLocalizationAlgorithm}. By contrast, our approach explicitly accounts for purposeful and potentially noncooperative interactions between multiple players by using iterative best response to search for local Nash equilibria of the players' variables.

\subsection{Multi-Player Path Planning via Dynamic Games}
\label{subsec: Multi-Player Path Planning via Dynamic Games}

In robotics applications, interactions between multiple players are naturally modeled as dynamic games. In particular, scenarios in which two groups of players have opposing objectives, such as robust control problems and pursuit-evasion games, are often formulated as zero-sum dynamic games \cite{fisac2015reach, FisacSastry2015PursuitEvasionDefense}. Meanwhile, problems in which multiple players have only partially conflicting objectives, such as path planning in busy traffic, are posed as general-sum dynamic games \cite{fridovich2019efficient,peters2021rss}. Although solutions to continuous-time dynamic games are characterized by coupled Hamilton-Jacobi-Bellman (HJB) PDEs \cite{starr1969nonzero, starr1969further, bansal2017hamilton}, solving these equations is typically intractable due to the so-called ``curse of dimensionality,'' \cite{bellman1966dynamic} i.e., their computation time grows exponentially in the state space dimension. For this reason, such methods are impractical in many multi-player scenarios of interest. 

In contrast, our work uses an iterative best response (IBR) scheme, in which each player takes a turn solving for their optimal strategy while assuming all other players' strategies are fixed \cite{fisac2018hierarchical, wang2018game, Kavuncu2021PotentialiLQR, ZanardiFrazzoli2021UrbanDrivingGamesWithLexicographicPreferences}. By replacing the dynamic game with a sequence of optimal control problems, the computational burden of solving for a local Nash equilibrium strategy is substantially reduced. Indeed, IBR has been successfully applied to a wide range of multi-player interaction scenarios, such as hierarchical planning for autonomous driving \cite{fisac2018hierarchical} and racing \cite{wang2018game}. Moreover, IBR is suitable for our application because it can be embedded in a factor graph-based framework, by iteratively solving estimation problems over the variables relevant to each player, while holding all other players' variables fixed.

Our work also draws inspiration from the potential games literature, which exploits the 
symmetric
cost structure of multi-player interactions in certain robotics applications \cite{Kavuncu2021PotentialiLQR, ZanardiFrazzoli2021UrbanDrivingGamesWithLexicographicPreferences}. 
Recent literature indicates that iterative methods which exploit this symmetry 
can be more efficient than those that do not \cite{Kavuncu2021PotentialiLQR}. Our approach draws inspiration from this observation: we recast the multi-agent SLAM problem under study as a potential game, and perform IBR in a manner that preserves its potential structure.


\section{PRELIMINARIES}
\label{sec: Preliminaries}

Below, we introduce core concepts in dynamic game theory. Readers are directed to \cite{basar1998DynamicNoncooperativeGameTheory} for further details.

\subsection{Open-Loop Nash Equilibria}

Consider the $N$-player, $K$-stage general-sum dynamic game $\Gamma$, with nonlinear, discrete-time system dynamics for each player $i \in [N] := \{1, \cdots, N\}$ and time $k \in [K] := \{1, \cdots, K\}$, given by:
\begin{equation} \label{Eqn: Dynamics, Player i}
    x_{k+1}^i \sim \mathcal{N} \big(f_k^i(x_k^i, u_k^i), \Sigma_f \big),
\end{equation}
where $x_k^i \in \R^{n_i}$, $u_k^i \in \R^{m_i}$, and $f_k^i: \R^{n_i} \times \R^{m_i} \ra \R^{n_i}$ are respectively the state, control input, and (differentiable) state transition map of player $i$ at time $k$, and $\Sigma_f$ is the associated covariance matrix.
Below, for each player $i \in [N]$, we use the shorthand $x^i := (x_1^i, \cdots, x_{K+1}^i) \in \R^{n_i(K+1)}$, and $u^i := (u_1^i, \cdots, u_K^i) \in \R^{m_i K}$. Moreover, we define $x := (x^1, \cdots, x^N) \in \R^{n(K+1)}$ and $u := (u^1, \cdots, u^N) \in \R^{mK}$, where $n = \sum_i n_i$ and $m = \sum_i m_i$.

In this work, we jointly estimate the trajectories of all players from the perspective of one particular player, referred to below as player 1 or the \textit{ego player}. Other players are termed \textit{non-ego players}. The ego player observes $N_\ell$ landmarks, whose global positions in $\R^{d_\ell}$ are given by $\ell := (\ell_1, \cdots, \ell_{N_\ell}) \in \R^{d_\ell N_\ell}$. Players also observe each others' positions. These measurements, at each time $k \in [K]$, are given by:
\begin{align}
    \landmarkmeas_k^\alpha &\sim \mathcal{N} \big(h(x_k^1, \ell_\alpha), \Sigma_h \big),~\textnormal{(landmark-agent)}\\
    \interplayermeas_k^{ij} &\sim \mathcal{N} \big(\overline{h}(x_k^i, x_k^j), \Sigma_{\overline{
    h}} \big),~\textnormal{(inter-agent)}
\end{align}
where $\landmarkmeas_k^\alpha \in \R^z$ is the measurement of landmark $\alpha$ by the ego player, for each $\alpha \in [N_\ell]$, while $\interplayermeas_k^{ij} \in \R^z$ is the measurement by player $i$ of player $j$, for each $i, j \in [N]$, $i \ne j$. Here, $\Sigma_h$ and $\Sigma_{\overline{h}}$ denote associated covariance matrices. 
Additionally, each player's objective is defined by $L^i(x, u)$, with $L^i: \R^{n(K+1)} \times \R^{mK} \ra \R$ for each $i \in [N], k \in [K]$. 
In this work, we presume that the ego player knows other players' objectives $L^i$. While this seems a strong assumption in practice, recent work has established that it is possible to infer unknown parameters of players' objectives in such games efficiently \cite{peters2021rss,cleac2020ral}.
Thus equipped, we now define the \textit{Nash equilibrium} of the GTP-SLAM problem.
\begin{definition} \label{Def: Open-Loop Nash Equilibrium}
(Open-Loop Nash equilibrium, \cite[Ch. 6]{basar1998DynamicNoncooperativeGameTheory})
We call $u^\star := \left(u^{1, \star}, \cdots, u^{N, \star} \right)$ an \emph{open-loop Nash equilibrium} of $\Gamma$ if no player can lower their cost by unilaterally deviating from their control $u^{i, \star}$ while all other players' controls, $u^{-i, \star}$, remains fixed, i.e.,
\begin{align}  \label{Eqn: Nash Equilibrium, Open-Loop}
    L^i\left(u^{i, \star}, u^{-i, \star} \right)
    \leq L^i\left(u^i, u^{-i, \star} \right), \hspace{5mm} \forall u^i \in \R^{m_i K}.
\end{align}
\end{definition}
\vspace{.2cm}

\subsection{Potential Dynamic Games}

Our approach leverages well-established convergence guarantees of iterative best response (IBR) algorithms in the setting of potential games. For clarity, we define a finite-stage potential game as follows.

\begin{definition}(Potential Dynamic Game, \cite{Kavuncu2021PotentialiLQR}, \cite{FonsecaMoralesHernandezLerma2018PotentialDifferentialGames}) \label{Def: Potential Dynamic Game}
An $N$-player, $K$-stage general-sum dynamic game $\Gamma$ is called a \emph{potential game} if there exists an optimal control problem, defined over all players' controls $(u^1, \cdots, u^N)$, whose solutions are Nash equilibria of the game $\Gamma$.
\end{definition}


In Section \ref{subsec: GTP-SLAM Potential Game}, we will recast the multi-player, noncooperative SLAM problem of interest into a potential game, and establish mild assumptions under which an appropriate IBR algorithm converges.

\section{METHODS}
\label{sec: Methods}

Our main contribution is GTP-SLAM, a novel SLAM algorithm for multi-player scenes, motivated by iterative best response. 
GTP-SLAM aims to jointly estimate the dynamic states and control inputs of all players in the scene, as well as landmark positions.
It does so from the ego player's perspective, while accounting for noncooperative, game-theoretic interactions between the players.


\subsection{Constructing the GTP-SLAM Factor Graph}
\label{subsec: Factor Graphs for Multi-Player Path Planning and SLAM}

\begin{figure*}[!htbp]
    \centering
    \includegraphics[scale=0.21]{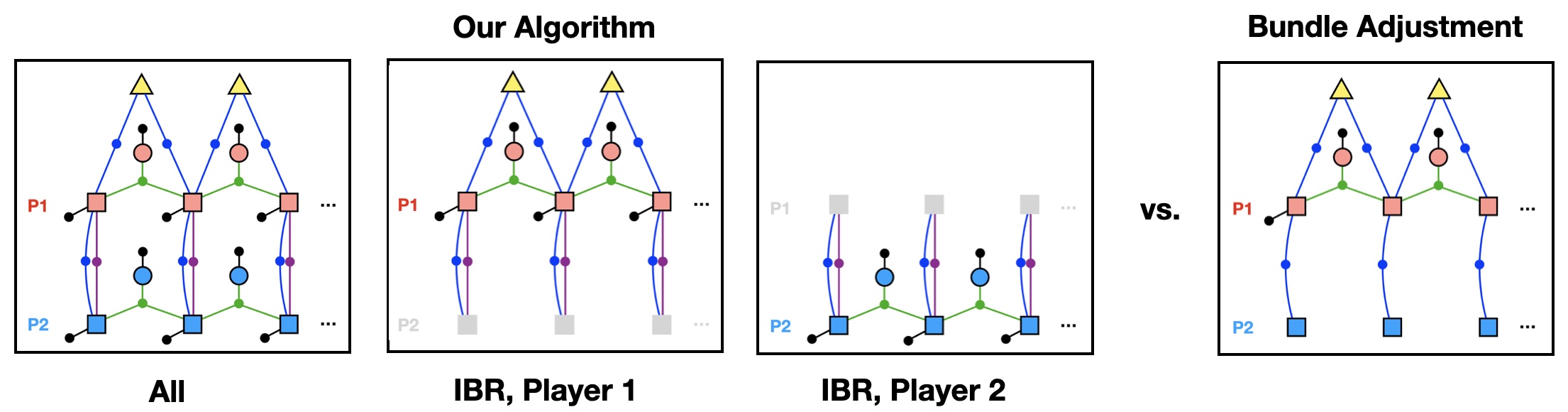}
    \caption{Factor 
    graphs for
    (Left) GTP-SLAM, our IBR-based algorithm and (Right) a standard bundle adjustment approach, for a two-player example. Red and blue nodes represent dynamic variables (states $x$, controls $u$) for players 1 and 2, respectively, while gray nodes indicate variables temporarily held constant. Square, circular, and triangular nodes represent states, controls, and landmarks, respectively. Green factors represent dynamics constraints, blue factors represent landmark and inter-player distance measurements, and black factors represent priors on states and controls, e.g. for lane tracking, heading alignment, and control effort minimization.
    }
    \label{fig:Factor_Graphs_GTSLAM}
\end{figure*}


We begin by expressing the players' noncooperative preferences as factors in a bipartite factor graph. 
Each factor is a function which encodes \emph{residual error} among the connected variables. 
That is, factors are vector-valued maps with which we may compute the joint likelihood of all input variables. Following standard Gaussian 
assumptions, we use the Mahalanobis distance associated to each factor (i.e., $\Vert v \Vert_\Sigma^2 := v^\top \Sigma^{-1} v$ for factor $v$ and covariance $\Sigma$) to compute the negative log-likelihood of a collection of variables.
Concretely, then, we construct a factor graph from the following terms:
\begin{align} \label{Eqn: Priors Factors}
    \text{priors}: \hspace{1mm} &g_k^i(x_k^i), \hat{g}_k^i(u_k^i) \\ \label{Eqn: Dynamics Factors}
    \text{dynamics}: \hspace{1mm} &f_k^i(x_k^i, u_k^i) - x_{k+1}^i \\ \label{Eqn: Interaction Factors}
    \text{interactions}: \hspace{1mm} &b(x_k^i, x_k^j) \\ \label{Eqn: Landmark Measurement Factors}
    \text{landmark measurements}: \hspace{1mm} &h(x_k^1, \ell_\alpha) - \landmarkmeas_k^\alpha \\ \label{Eqn: Inter-Player Measurement Factors}
    \text{inter-player observations}: \hspace{1mm} &\overline{h}(x_k^i, x_k^j) - \interplayermeas_k^{ij},
\end{align}
which are color-coded in Figure~\ref{fig:Factor_Graphs_GTSLAM}.
For example, the ternary dynamics factor \eqref{Eqn: Dynamics Factors} computes the difference between vehicles' states and those predicted by the appropriate state transition function \eqref{Eqn: Dynamics, Player i}.
Likewise, the factors \eqref{Eqn: Interaction Factors} between pairs of states belonging to players $i \ne j$ describe interactions between pairs of players.
For example, to encode collision avoidance, 
we may set $b(x, x') := 1/\Vert x - x' \Vert_2$.
The factors $h(x_k^1, \ell_\alpha) - \landmarkmeas_k^\alpha$ denote the difference between expected and actual landmark measurements made by the ego player.
Finally, $\overline{h}(x_k^i, x_k^j) - \interplayermeas_k^{ij}$, where $i, j \in [N], i \ne j, \min\{i, j\} = 1$, denotes inter-player position measurements between the ego player and each non-ego player.

The maximum a posteriori (MAP) estimation problem faced by each player, then, is the minimization of a sum of squared factors. In other words, each player's individual decision problem is a nonlinear least squares problem, when other players' variables are held constant.
Neglecting interaction factors, landmarks, and inter-player measurements (which couple players' variables together), 
we compute the partial log-likelihood of player $i$'s variables as: 
\begin{align} \label{Eqn: Player i, cost, bare}
    J^i(x^i, u^i) &:= \sum_{k=1}^K \Big( \Vert g_k^i(x_k^i) \Vert_{\Sigma_g}^2 
    + \Vert \hat{g}_k^i(u_k^i) \Vert_{\Sigma_{\hat{g}}}^2 \\ \nonumber
    &\hspace{1cm} + \Vert f_k^i(x_k^i, u_k^i) - x_{k+1}^i \Vert_{\Sigma_f}^2\Big).
\end{align}
Note that \eqref{Eqn: Player i, cost, bare} does not include interaction terms ($b$) or measurements ($\landmarkmeas$ or $\interplayermeas$). This is because these quantities depend upon multiple players' variables jointly, and also because measurements pertaining to landmarks and other players' states are only assumed to be collected by the ego player. Including these terms, 
the ego player's full estimation problem is thus given by:
\begin{align} \label{Eqn: L1}
    L^1(x, u, \ell) &:= J^1(x, u) + \sum_{\alpha=1}^{N_{\ell}} \sum_{k=1}^K \Vert h(x_k^1, \ell_\alpha) - \landmarkmeas_k^\alpha \Vert_{\Sigma_h}^2\nonumber \\
    &\hspace{1cm} + \sum_{j \ne 1} \sum_{k=1}^K \Vert \overline{h}(x_k^1, x_k^j) - \interplayermeas_k^{1j} \Vert_{\Sigma_{\overline{h}}}^2 \nonumber\\
    &\hspace{1cm} + \sum_{j \ne 1} \sum_{k=1}^K \Vert b(x_k^1, x_k^j) \Vert_{\Sigma_b}^2 ,
\end{align}
while each non-ego player's MAP problem is given by:
\begin{align} \label{Eqn: Li}
    L^i(x, u) &:= J^i(x^i, u^i) + \sum_{k=1}^K \Vert \overline{h}(x_k^i, x_k^1) - \interplayermeas_k^{i1} \Vert_{\Sigma_{\overline{h}}}^2\nonumber \\
    &\hspace{1cm} + \sum_{j \ne i} \sum_{k=1}^K \Vert b(x_k^i, x_k^j) \Vert_{\Sigma_b}^2,
\end{align}
for each $i \in [N] \backslash \{1\}$.

\subsection{GTP-SLAM as a Potential Game}
\label{subsec: GTP-SLAM Potential Game}

Next, we illustrate that the GTP-SLAM problem of Section~\ref{subsec: Factor Graphs for Multi-Player Path Planning and SLAM} is a potential game  (Lemma~\ref{lem: Potential Structure}, Proposition~\ref{Prop: Potential Game}).
This connection to potential games is critical, as it suggests a locally-convergent solution method for GTP-SLAM problems given in Section~\ref{subsec: Iterative Best Response} (Corollary~\ref{cor: IBR Converges}). 
The following results are based upon established concepts in the literature \cite{FonsecaMoralesHernandezLerma2018PotentialDifferentialGames, ZanardiFrazzoli2021UrbanDrivingGamesWithLexicographicPreferences, Kavuncu2021PotentialiLQR}; here, we illustrate their pertinence to the noncooperative SLAM problem. 


\begin{lemma} \label{lem: Potential Structure}
Consider an $N$-player, $K$-stage dynamic game $\Gamma$, with fixed initial condition $x_1 := (x_1^1, \cdots, x_1^N) \in \R^n$. Suppose the system dynamics of each player given by \eqref{Eqn: Dynamics, Player i}, and the cost function of each player $i$ is of the form:
\begin{align*}
    L^1(x, u, \ell) &:= C^1(x^1, u^1, \ell) + \sum_{j=2}^N C^{1j}(x^1, x^j), \\
    L^i(x, u) &:= C^i(x^i, u^i) + \sum_{j \ne i} C^{ij}(x^i, x^j), 
    \hspace{2mm} \forall i \in [N] \backslash \{1\}
\end{align*}
respectively, where $C^i: \R^{n_i (K+1)} \times \R^{m_i K} \ra \R$, and $C^{ij}: \R^{n_i (K+1)} \times \R^{n_j (K+1)} \ra \R$ satisfy:
\begin{align*}
    C^{ij}(x^i, x^j) = C^{ji}(x^j, x^i), \hspace{5mm} \forall \hspace{0.5mm} i, j \in [N], i \ne j.
\end{align*}
Then $\Gamma$ is a potential game corresponding to the optimal control problem of minimizing the potential function:
\begin{align} \label{Eqn: Potential Function, p(x, u)}
    p(x, u) := \sum_{i=1}^N C^i(x^i, u^i) + \sum_{i, j \in [N], i < j} C^{ij}(x^i, x^j),
\end{align}
subject to the dynamics \eqref{Eqn: Dynamics, Player i}, for each $i \in [N]$. 

\begin{proof}
Refer to the appendix.
\end{proof}
\end{lemma}

Given the result of Lemma~\ref{lem: Potential Structure}, we now show that the GTP-SLAM game structure given in \eqref{Eqn: Player i, cost, bare} is consistent with a potential game.

\begin{proposition} \label{Prop: Potential Game}
The GTP-SLAM game, with players' objectives given by \eqref{Eqn: Player i, cost, bare}, is a potential game.

\begin{proof}
In the context of \eqref{Eqn: Player i, cost, bare}, we have:
\begin{align*}
    C^1(x^1, u^1, \ell) &:= J^1(x^1, u^1) + \sum_{\alpha=1}^{N_{\ell}} \sum_{k=1}^K \Vert h(x_k^1, \ell_\alpha) - \landmarkmeas_k^\alpha \Vert_{\Sigma_h}^2, \\
    C^{1j}(x^1, x^j) &:= \sum_{k=1}^K \Vert \overline{h}(x_k^1, x_k^j) - \interplayermeas_k^{1j} \Vert_{\Sigma_{\overline{h}}}^2 \\
    &\hspace{5mm} + \sum_{k=1}^K \Vert b(x_k^1, x_k^j) \Vert_{\Sigma_b}^2, \hspace{2mm} \forall \hspace{0.5mm} j \in [N] \backslash \{1\}, \\
    C^i(x^i, u^i) &:= J^i(x^i, u^i), \hspace{2mm} \forall \hspace{0.5mm} i \in [N] \backslash \{1\}, \\
    C^{i1}(x^i, x^1) &:= \sum_{k=1}^K \Vert \overline{h}(x_k^i, x_k^1) - \interplayermeas_k^{i1} \Vert_{\Sigma_{\overline{h}}}^2 \\
    &\hspace{5mm} + \Vert b(x_k^i, x_k^1) \Vert_{\Sigma_b}^2 \forall \hspace{0.5mm} i \in [N] \backslash \{1\}, \\
    C^{ij}(x^i, x^j) &:=  \sum_{k=1}^K \Vert b(x_k^i, x_k^j) \Vert_{\Sigma_b}^2, \hspace{2mm} \forall \hspace{0.5mm} i, j \in [N] \backslash \{1\}, i \ne j.
\end{align*}
Thus, by Lemma~\ref{lem: Potential Structure}, the game encoded in GTP-SLAM is a potential dynamic game. 
\end{proof}
\end{proposition}


\subsection{Iterative Best Response}
\label{subsec: Iterative Best Response}


To find Nash equilibria of the GTP-SLAM game with objectives given by \eqref{Eqn: L1} and \eqref{Eqn: Li}, we employ Algorithm \ref{Alg: GTP-SLAM}, an approach inspired by iterative best response (IBR). 
Specifically, Algorithm~\ref{Alg: GTP-SLAM} proceeds in rounds, where each player $i$ minimizes its MAP objective $L^i$ while holding variables pertaining to other players $j \ne i$ fixed. 
Convergence is guaranteed by the following corollary due to \cite{ZanardiFrazzoli2021UrbanDrivingGamesWithLexicographicPreferences}.

\begin{corollary} \label{cor: IBR Converges}
Algorithm \ref{Alg: GTP-SLAM} converges to an open-loop Nash equilibrium when applied to potential games of the form of Definition \ref{Def: Potential Dynamic Game}, if the maximum number of iterations $Q$ is set to $\infty$ and the convergence tolerance $\epsilon$ is set to $0$.
\end{corollary}

\begin{proof}
See \cite[Proposition 1]{ZanardiFrazzoli2021UrbanDrivingGamesWithLexicographicPreferences}
\end{proof}

\begin{algorithm} 
{
\small
\SetAlgoLined


\KwData{Maximum number of iterations $Q$, Convergence tolerance $\epsilon$, Cost functions $L^i$.} 
\KwResult{Nash equilibrium variables $x^\star = (x^{1,\star}, \cdots, x^{N, \star}), u^\star = (u^{1,\star}, \cdots, u^{N, \star}), \ell^\star$.}

 \vspace{2mm}
 
 $q \gets 1$
 
 \While{$q \leq Q$ \emph{\textbf{and}} $\Vert x^{i, q} - x^{i, q-1} \Vert_2 < \epsilon, \hspace{0.5mm} \forall \hspace{0.5mm} i \in [N]$ }{
 
 $(x^{1, q}, u^{1, q}, \ell^q) \leftarrow \text{arg}  \min\limits_{(x^1, u^1, \ell)} 
    L^1(x, u, \ell)$
 
 \For{$i \in [N]\backslash\{1\}$}{
    $(x^{i, q}, u^{i, q}) \leftarrow \text{arg}  \min\limits_{(x^i, u^i)} 
    L^i(x, u)$
 }

 $q \gets q + 1$
 }
 
 \Return{$(x^{1,q}, \cdots, x^{N, q}), (u^{1,q}, \cdots, u^{N, q}), \ell^q$}
 \caption{Solving the GTP-SLAM Problem}
 \label{Alg: GTP-SLAM}
 }
\end{algorithm}

\section{RESULTS}
\label{sec: Results}

\begin{figure*}[!htbp]
    \centering
    \includegraphics[scale=0.2]{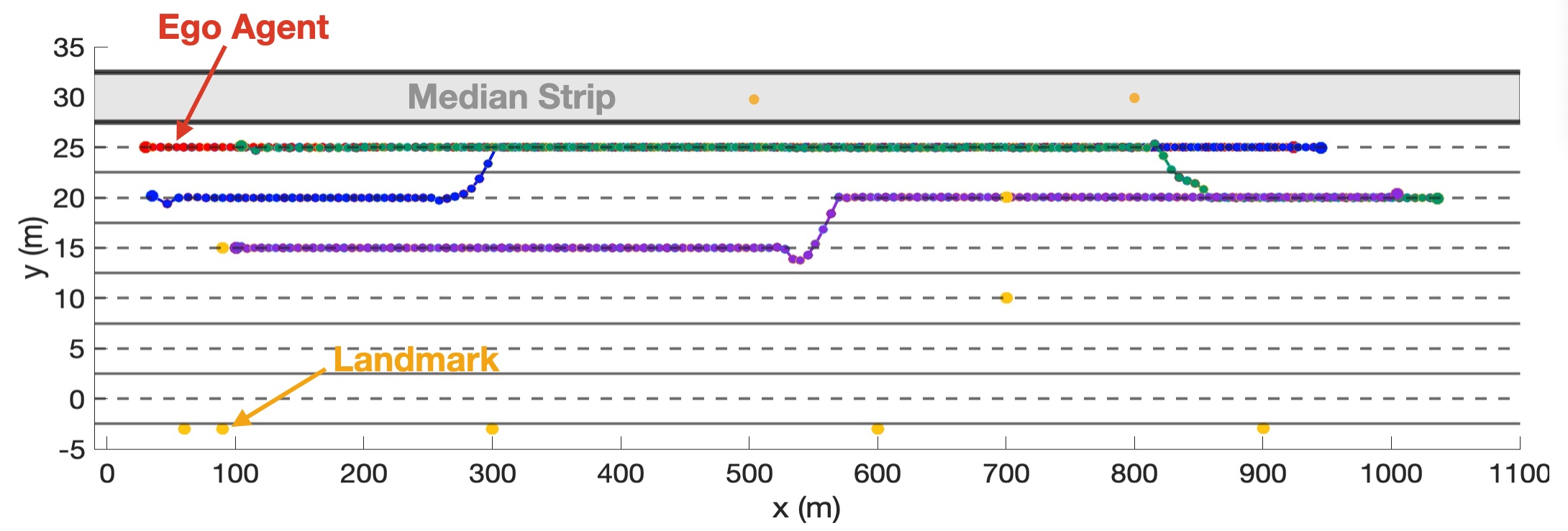}
    \caption{Schematic of the highway example. Here, players 1 (red), 2 (blue), 3 (green), and 4 (purple) navigate a kilometer-long stretch of highway and interact with each other while performing lane changes.
    The ego player detects landmarks in the scene, which describe objects common to realistic highway scenarios, e.g., speed limit signs, exit signs, light poles, etc.
    }
    \label{fig:Highway_Example_Schematic___Annotated}
\end{figure*}

\subsection{Simulation Setup}
\label{subsec: Simulation Setup}

To demonstrate the importance of game-theoretic priors in multi-player SLAM problems, we simulate a highway driving scenario (Figure \ref{fig:Highway_Example_Schematic___Annotated}). 
Specifically, four vehicles change lanes over a kilometer-long stretch of highway while avoiding collision, maintaining a desired speed, and collecting range and bearing measurements of surrounding landmarks via lidar. 
We assume that each vehicle follows Dubins paths, i.e., moves with constant speed ($\SI{30}{\meter\per\second}$, here), and can control its yaw rate.
Vehicle motion is discretized at intervals of $\SI{0.2}{\second}$.
These dynamics constitute the state transition maps $f_k^i$.
Moreover, we assume that the highway is sparsely populated with occasional landmarks, e.g., exit signs, speed limit signs, and light poles, shown as yellow circles in Figure~\ref{fig:Highway_Example_Schematic___Annotated}.

A local Nash equilibrium of the highway driving game is found by applying Algorithm~\ref{Alg: GTP-SLAM} with fixed initial states for all players and neglecting measurement likelihood factors. 
To understand the role of game-theoretic interactions in SLAM problems, we conduct a Monte Carlo study of the highway driving scenario of Figure~\ref{fig:Highway_Example_Schematic___Annotated}, with results recorded in Figure~\ref{fig:Error_Plots___Quartiles___IBR_Order_2341}.
For each noise standard deviation level in the set $\{0.05, 0.10, \cdots, 0.95, 1.0\}$, we ran 50 experiments, each with a slightly perturbed set of initial conditions. For each experiment, we simulated random measurements of all landmarks, and of all non-ego players' planar coordinates, with respect to the ego player's local frame. 
We then ran Algorithm~\ref{Alg: GTP-SLAM} to convergence, and compared the results to a standard bundle adjustment approach that neglected game-theoretic priors.
That is, by ignoring \eqref{Eqn: Interaction Factors} for the ego player and \eqref{Eqn: Priors Factors}, \eqref{Eqn: Dynamics Factors}, \eqref{Eqn: Interaction Factors} for all non-ego players, the GTP-SLAM problem reduces to a single MAP problem which may be solved jointly for all players at once. Throughout all simulations, we use GTSAM \cite{dellaert2012gtsam} to construct the factors above, compute Jacobians, and implement Levenberg-Marquardt steps \cite{nocedal2006numerical} for both GTP-SLAM and bundle adjustment.





\subsection{Discussion}
\label{subsec: Discussion}

\begin{figure}[ht]
    \centering
    \includegraphics[scale=0.12]{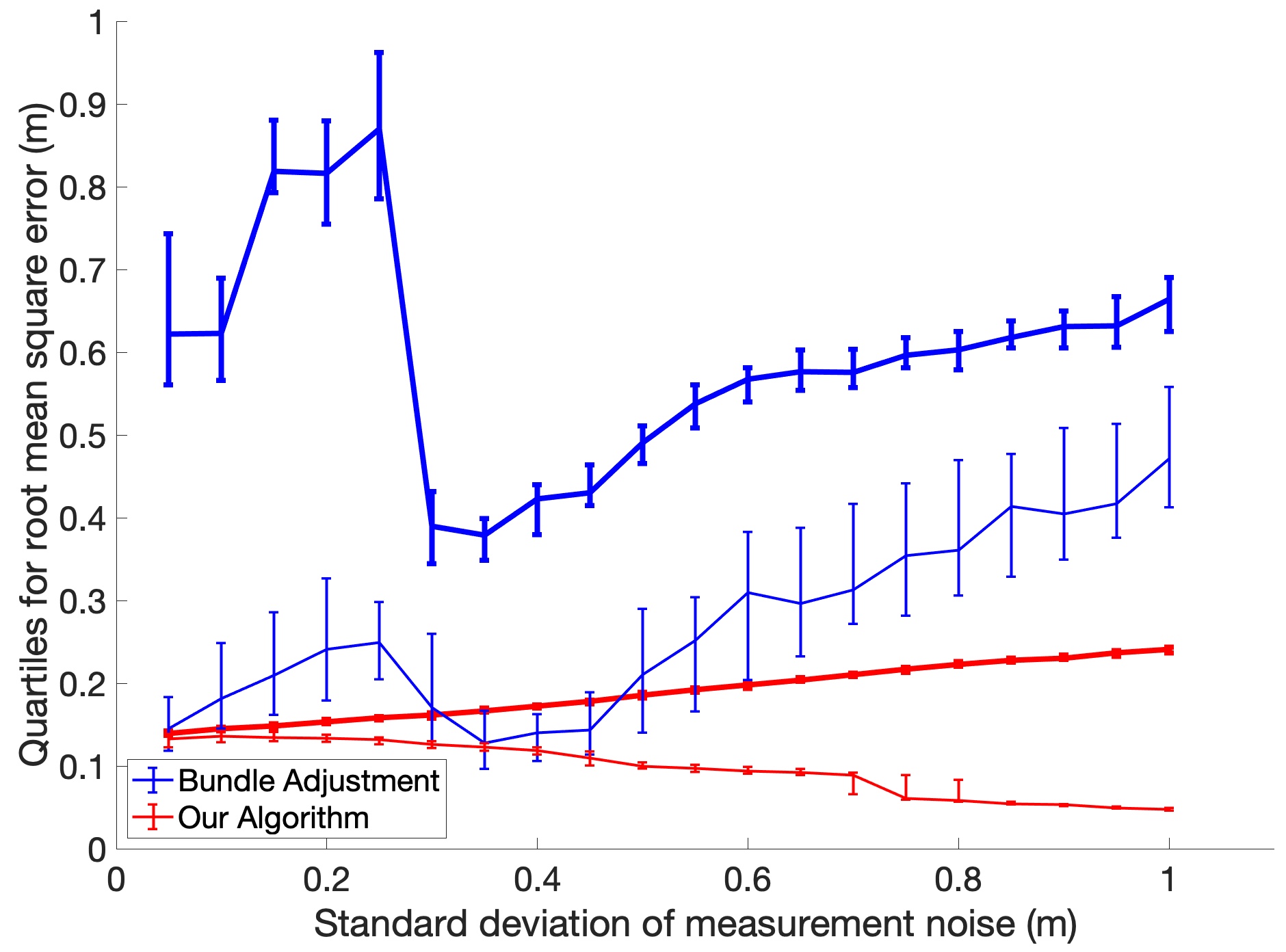}
    \caption{Root-mean-square error of estimated vehicle (thick) and landmark (thin) positions vs. standard deviation of measurement noise, for player positions and landmarks.
    }
    \label{fig:Error_Plots___Quartiles___IBR_Order_2341}
\end{figure}

Figure~\ref{fig:Error_Plots___Quartiles___IBR_Order_2341} records the localization and map reconstruction error of GTP-SLAM (red) and standard bundle adjustment (blue). Compared to the bundle adjustment baseline, the localization and map reconstruction error for GTP-SLAM is lower across all noise standard deviation levels, and degrades more gracefully as noise levels increase. 
In particular, conventional bundle adjustment becomes numerically unstable at low noise levels; by contrast, the introduction of game-theoretic priors appears to yield a more well-conditioned estimation problem.
In summary, these results indicate that game-theoretic priors introduce additional structure in an otherwise complex estimation problem, enabling reliable recovery of vehicle states and map landmarks.



\section{CONCLUSION AND FUTURE WORK}
\label{sec: Conclusion and Future Work}

Inspired by recently developed iterative dynamic game algorithms, we present a novel method for Simultaneous Localization and Mapping (SLAM) in dynamic scenes in which multiple players interact noncooperatively. Our approach 
exploits the structure of potential games to ensure reliable convergence. 
Empirical results illustrate that our algorithm outperforms standard bundle adjustment methods in localization and map reconstruction accuracy. 

We foresee several important directions for future work. First, 
our experiments do not yet consider loop closures, which are essential for the long-term recovery of static scenes in SLAM tasks. It is thus critical to study how to best incorporate game-theoretic priors when detecting and enforcing loop closures.
Second, we demonstrated our method in full-graph optimization problems; in practice, however, SLAM graphs are often optimized incrementally, as measurements are acquired in real-time. Our approach readily extends to this setting.
Finally, our method only computes open-loop game strategies, corresponding to feedforward, rather than feedback, controls. 
Future work will investigate game-theoretic SLAM priors in more complicated strategy spaces.








\printbibliography

\appendix[Proof of Proposition~\ref{Prop: Potential Game}]
\label{sec: Appendix}

Proposition \ref{Prop: Potential Game} follows directly from analogous proofs established in \cite{Kavuncu2021PotentialiLQR, GonzalezSanchez2014DynamicPotentialGames}, rephrased here for completeness.

\begin{proof}
Let $p(x, u)$ be given by \eqref{Eqn: Potential Function, p(x, u)}, with a minimizer given by $u^\star := (u^{1, \star}, \cdots, u^{N, \star}) \in \R^{mK}$, $x^\star := (x^{1, \star}, \cdots, x^{N, \star}) \in \R^{n(K+1)}$. For any player $r \in [N]$, and any unilateral deviation in player $r$'s controls away from $u^\star$, i.e., $u := (u^r, u^{-r, \star}) = (u^{1, \star}, \cdots, u^{r-1, \star}, u^{r}, u^{r+1, \star}, \cdots, u^{N, \star}) \in \R^{mK}$, with corresponding state trajectory $(x^r, x^{-r, \star}) = (x^{1, \star}, \cdots, x^{r-1, \star}, x^r, x^{r+1, \star}, \cdots, x^{N, \star}) \in \R^{n(K+1)}$:
\begin{align*}
    0 &\leq p(x^{r}, x^{-r, \star}, u^{r}, u^{-r, \star}) - p(x^{r, \star}, x^{-r, \star}, u^{r, \star}, u^{-r, \star}) \\
    &= \Bigg( C^{r}(x^{r}, u^{r}) + \sum_{i \ne r} C^{ir}(x^{i, \star}, x^{r}) \\
    &\hspace{1cm} + \sum_{i \ne r} C^i(x^{i, \star}, u^{i, \star}) + \sum_{\stackrel{1 \leq i < j \leq n}{i, j \ne r}} C^{ij}(x^{i, \star}, x^{j, \star}) \Bigg) \\
    &\hspace{3mm} - \Bigg( C^{r}(x^{r, \star}, u^{r, \star}) + \sum_{i \ne r} C^{ir}(x^{i, \star}, x^{r, \star}) \\
    &\hspace{5mm} + \sum_{i \ne r} C^i(x^{i, \star}, u^{i, \star}) + \sum_{\stackrel{1 \leq i < j \leq n}{i, j \ne r}} C^{ij}(x^{i, \star}, x^{j, \star}) \Bigg) \\
    &= \Bigg( C^{r}(x^{r}, u^{r}) + \sum_{i \ne r} C^{ir}(x^{i, \star}, x^{r}) \Bigg) \\
    &\hspace{5mm} - \Bigg( C^{r}(x^{r, \star}, u^{r, \star}) + \sum_{i \ne r} C^{ir}(x^{i, \star}, x^{r, \star}) \Bigg) \\
    &= L^r(x^r, x^{-r, \star}, u^{r}, u^{-r, \star}) - L^r(x^\star, u^\star).
\end{align*}
(We have made implicit the dependence of $L^1$ and $C^1$ on the landmarks $\ell$, for notational convenience.)
This condition matches that which defines open-loop Nash equilibria (Definition~\ref{Def: Open-Loop Nash Equilibrium}). Hence, $u^\star$ is an open-loop Nash equilibrium of the game $\Gamma$.
\end{proof}





\end{document}